\documentclass{ieeeaccess}
\usepackage{cite}
\usepackage{url}
\usepackage{amsmath,amssymb,amsfonts}
\usepackage{algorithmic}
\usepackage{graphicx}
\usepackage{textcomp}
\usepackage{comment}
\usepackage[utf8]{inputenc}
\PassOptionsToPackage{hyphens}{url}\usepackage[hidelinks]{hyperref}
\pdfoutput=1

\usepackage{bm}
\usepackage{listings}
\makeatletter
\AtBeginDocument{\DeclareMathVersion{bold}
\SetSymbolFont{operators}{bold}{T1}{times}{b}{n}
\SetSymbolFont{NewLetters}{bold}{T1}{times}{b}{it}
\SetMathAlphabet{\mathrm}{bold}{T1}{times}{b}{n}
\SetMathAlphabet{\mathit}{bold}{T1}{times}{b}{it}
\SetMathAlphabet{\mathbf}{bold}{T1}{times}{b}{n}
\SetMathAlphabet{\mathtt}{bold}{OT1}{pcr}{b}{n}
\SetSymbolFont{symbols}{bold}{OMS}{cmsy}{b}{n}
\renewcommand\boldmath{\@nomath\boldmath\mathversion{bold}}}
\makeatother

\def\BibTeX{{\rm B\kern-.05em{\sc i\kern-.025em b}\kern-.08em
    T\kern-.1667em\lower.7ex\hbox{E}\kern-.125emX}}

\usepackage{colortbl}
\usepackage{soul}

\begin{document}

\clearpage
\onecolumn
\thispagestyle{empty}

\begin{center}
{\Large\bfseries Notice}\par\vspace{1.25em}

This manuscript has been \textbf{accepted and published} in \textit{IEEE Access}.\\
Please cite and consult the IEEE Version of the manuscript.\par\vspace{0.8em}

\textbf{BibTeX}
\end{center}
\begin{lstlisting}
@ARTICLE{Kuzman2025LLMTeacherStudent,
  author={Kuzman, Taja and Ljube{\v{s}}i{\'c}, Nikola},
  journal={IEEE Access}, 
  title={LLM Teacher-Student Framework for Text Classification With No Manually Annotated Data: A Case Study in IPTC News Topic Classification}, 
  year={2025},
  volume={13},
  number={},
  pages={35621-35633},
  keywords={Data models;Annotations;Media;Manuals;Multilingual;Computational modeling;Training;Training data;Transformers;Text categorization;Multilingual text classification;IPTC;large language models;LLMs;news topic;topic classification;training data preparation;data annotation},
  doi={10.1109/ACCESS.2025.3544814},
  url={https://ieeexplore.ieee.org/document/10900365}
  }
\end{lstlisting}

\vspace{2em}

\textbf{Note for arXiv readers.} This arXiv version is retained for
archival purposes. Readers should use and cite the \textit{IEEE Access}
Version available at \url{https://ieeexplore.ieee.org/document/10900365}
\vspace{1.5em}

\textbf{Open Access License.} The published article is available under the
Creative Commons Attribution 4.0 International (CC BY 4.0) license. You may
share and adapt the material for any purpose, even commercially, provided
appropriate credit is given to the original source. See
\href{https://creativecommons.org/licenses/by/4.0/}{creativecommons.org/licenses/by/4.0/}.

\vspace*{\fill}
\clearpage
\twocolumn
\setcounter{page}{1}

\history{Date of publication 24 February, 2025, date of current version 20 February, 2025.}
\doi{10.1109/ACCESS.2025.3544814}

\title{LLM Teacher-Student Framework for Text Classification With No Manually Annotated Data: A Case Study in IPTC News Topic Classification}
\author{\uppercase{Taja Kuzman}\authorrefmark{1,2}, 
\uppercase{Nikola Ljube{\v{s}}i{\'c}}\authorrefmark{1,3}
}

\address[1]{Department of Knowledge Technologies, Jožef Stefan Institute, 1000 Ljubljana, Slovenia (e-mail: \{taja.kuzman,nikola.ljubesic\}@ijs.si)}
\address[2]{Jožef Stefan International Postgraduate School, 1000 Ljubljana, Slovenia}
\address[3]{University of Ljubljana, 1000 Ljubljana, Slovenia}
\tfootnote{This work was supported by the projects ``Embeddings-based techniques for Media Monitoring Applications'' (L2-50070, co-funded by the Kliping d.o.o. agency), ``Large Language Models for Digital Humanities'' (GC-0002), and the research programme ``Language resources and technologies for Slovene'' (P6-0411), all funded by the Slovenian Research and Innovation Agency (ARIS).}

\markboth
{Kuzman and Ljubešić: LLM Teacher-Student Framework for IPTC News Topic Classification}
{Kuzman and Ljubešić: LLM Teacher-Student Framework for IPTC News Topic Classification}

\corresp{Corresponding author: Taja Kuzman (e-mail: taja.kuzman@ijs.si).}

\begin{abstract}
With the ever-increasing number of news stories available online, classifying them by topic, regardless of the language they are written in, has become crucial for enhancing readers' access to relevant content. To address this challenge, we propose a teacher-student framework based on large language models (LLMs) for developing multilingual news topic classification models of reasonable size with no need for manual data annotation. The framework employs a Generative Pretrained Transformer (GPT) model as the teacher model to develop a news topic training dataset through automatic annotation of 20,000 news articles in Slovenian, Croatian, Greek, and Catalan. Articles are classified into 17 main categories from the Media Topic schema, developed by the International Press Telecommunications Council (IPTC). The teacher model exhibits high zero-shot performance in all four languages. Its agreement with human annotators is comparable to that between the human annotators themselves. To mitigate the computational limitations associated with the requirement of processing millions of texts daily, smaller BERT-like student models are fine-tuned on the GPT-annotated dataset. These student models achieve high performance comparable to the teacher model. Furthermore, we explore the impact of the training data size on the performance of the student models and investigate their monolingual, multilingual, and zero-shot cross-lingual capabilities. The findings indicate that student models can achieve high performance with a relatively small number of training instances, and demonstrate strong zero-shot cross-lingual abilities. Finally, we publish the best-performing news topic classifier, enabling multilingual classification with the top-level categories of the IPTC Media Topic schema.
\end{abstract}


\titlepgskip=-21pt

\maketitle

\section{Introduction}
\label{sec:introduction}

\PARstart{T}{opic} classification is a significant asset in the news industry, enabling automatic identification of news topics and facilitating readers' access to content that aligns with their interests. However, developing a robust news topic classifier is challenging, particularly because of the absence of manually annotated data, which are especially scarce for non-English languages. Manual annotation, although invaluable, is an expensive and labor-intensive process that cannot be quickly extended to numerous languages.

\Figure[t!](topskip=0pt, botskip=0pt, midskip=0pt)[width=0.9\textwidth]{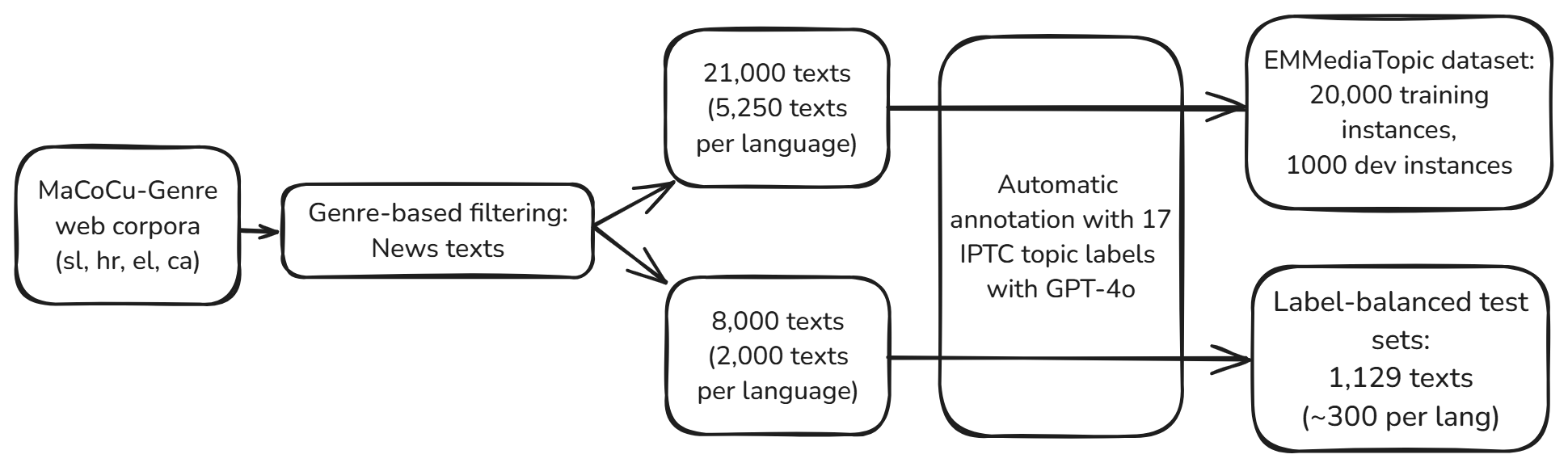}
{ \textbf{Pipeline for the preparation of training and test datasets for the classification of news topics.}\label{fig:dataset-pipeline}}

Despite the emergence of zero-shot approaches that employ Generative Pretrained Transformers (GPTs) that have demonstrated impressive results in various natural language processing tasks in diverse languages \cite{kuzman2023automatic}, \cite{wibowo2024copal,chowdhery2023palm}, their application in settings with extensive data to be processed remains impractical due to their high computational demands. A more suitable technology for this setting is a smaller BERT-like language model; however, such models require thousands of annotated instances for effective fine-tuning for a certain task.

In this paper, we introduce an approach to developing annotated training data and multilingual BERT-like news topic classifiers through a teacher-student framework based on large language models (LLMs). Our methodology leverages the power of GPTs that we use as teacher models to automatically annotate news articles with top-level Media Topic labels, introduced by the International Press Telecommunications Council (IPTC) \cite{iptcmediatopics}. This automatically annotated dataset is then used to train a smaller XLM-RoBERTa-based student model, addressing the computational demands of the news media industry, where topic annotation needs to be scaled to millions of data points.

In our experiments, we train and evaluate the models on multilingual training and test datasets comprising four diverse languages: Catalan, Croatian, Greek, and Slovenian. The selection of these languages for evaluation is based on their availability in the MaCoCu corpora collection~\cite{banon2022macocu} to ensure a high level of comparability between language-specific datasets. 
Furthermore, these languages exhibit varying degrees of linguistic relatedness, with Croatian and Slovenian being closely related, and the other two languages belonging to separate branches of the Indo-European language family. This distinction provides valuable insights for cross-lingual experiments.


In this work, our aim is to address the following research questions, pertaining to the task of IPTC news topic classification:
\begin{enumerate}
 \item[(RQ1)] Can the teacher LLM, used as a data annotator for news topic classification, achieve annotation quality comparable to that of human annotators?
\item[(RQ2)] How much data, annotated by the teacher model, are required for the student model to achieve performance comparable to the teacher model?
\item[(RQ3)] Is it necessary to include the target language in the training data, or does the fine-tuned student model exhibit satisfactory zero-shot cross-lingual capabilities?
\item[(RQ4)] Does fine-tuning a student model on target-language monolingual data yield better outcomes than training on a multilingual dataset of equal size?
\end{enumerate}

In summary, this study makes the following contributions. We investigate the feasibility of the LLM teacher-student framework for the development of accurate and computationally efficient multilingual news topic classifiers for languages that do not have readily accessible manually annotated training data suitable for the task. Specifically, we focus on single-label multi-class text classification employing top-level categories of the IPTC NewsCodes Media Topic schema. This study assesses the applicability of a GPT model for data annotation by comparing the level of agreement between the model and human annotators to the level of agreement among the human annotators themselves. Furthermore, we examine the classification performance of the student model relative to its teacher model, while also exploring the impact of training data size on the student model's performance, as well as its zero-shot cross-lingual and multilingual capabilities. Motivated by these positive results, we publish the best performing model in the Hugging Face repository (\url{https://huggingface.co/classla/multilingual-IPTC-news-topic-classifier}), providing an IPTC news topic classification model that can be applied to any of the 100 languages covered by the XLM-RoBERTa model. To the best of our knowledge, this is the first openly-available multilingual topic classifier that uses the (top-level) IPTC Media Topic categories.

The remainder of this paper is organized as follows. Section \ref{sec:related-work} provides an overview of the existing literature on IPTC news topic classification and the application of GPT models to data annotation. In Section \ref{sec:training-data}, we introduce our methodology for automatic annotation of training data by leveraging the zero-shot capabilities of large language models. In Section \ref{sec:test-data}, we describe the manual annotation campaign that was undertaken to develop a test set for evaluating the performance of both teacher and student models, and we evaluate the performance of the GPT model as a data annotator compared to human annotators. Section \ref{sec:experiments} presents the experiments that involve fine-tuning BERT-like models on varying sizes of training data, and the analysis of their monolingual, multilingual, and zero-shot cross-lingual performance. Section \ref{sec:conclusions} concludes the paper with a discussion of the main findings and suggestions for future research. Additionally, the Appendix provides more details on the annotation guidelines and the description of the top-level IPTC Media Topic labels (Section \ref{sec:annotation-guidelines}), and the prompt used for automatic data annotation (Section \ref{sec:app-prompt}).

\section{Related Work}
\label{sec:related-work}

The International Press Telecommunications Council (IPTC) is a global organization dedicated to the development and promotion of industry standards for the exchange of news data. Among its notable contributions is the maintenance of IPTC NewsCodes controlled vocabularies that are widely adopted by major media providers, including Agence France-Presse, Associated Press, and Reuters \cite{iptcNewsCodesIPTC}. The NewsCodes vocabularies include the Media Topic vocabulary, which has been established as a global standard to ensure consistent coding of news metadata across diverse news providers \cite{iptcWhatIPTC}. The Media Topic taxonomy is structured hierarchically, with 17 topic labels at the top level. The entire taxonomy encompasses more than 1,000 topic labels, organized across up to five levels of granularity \cite{iptcGroupsNewsCodes}.

Multiple previous studies investigated news topic classification \cite{roberts2021media,chy2014bangla}, \cite{pranjic2020evaluation,misra2022news}, \cite{kosem2023spremljevalni}. However, there is no manually annotated reference dataset that can be used directly for the training and evaluation of IPTC Media Topic classifiers. The existing topic datasets have several limitations:
\begin{enumerate}
    \item \textbf{Inconsistent Topic Schemata:} Many datasets use their own topic schemata \cite{misra2022news,roberts2021media,kosem2023spremljevalni}, rendering the results of their experiments highly dataset dependent and consequently incomparable with other studies.
    \item \textbf{Lack of Manual Annotation:} Many topic datasets did not undergo manual annotation. Instead, topic categories were derived from source media websites \cite{misra2022news,kosem2023spremljevalni} or automatically annotated using topic modeling \cite{roberts2021media} based on the word2vec algorithm \cite{mikolov2013distributed}.
    \item \textbf{Use of a Deprecated IPTC Schema:} Some datasets use the outdated IPTC Subject Codes schema (see, for instance, \cite{stt-fi-1992} and \cite{pranjic2020evaluation}), which was replaced by the IPTC Media Topic schema in 2010. Additionally, the IPTC Media Topic schema itself is subject to frequent updates, occurring every two to three months \cite{iptcNewsCodesGuidelines}, which poses additional challenges for the development of a stable reference dataset and classifiers.
\end{enumerate}

Recently, two news topic datasets have been introduced, both annotated with the IPTC Media Topic schema: the MN-DS dataset \cite{petukhova2023mn}, comprising approximately 10,000 English news articles, and the EventDNA dataset \cite{colruyt2023eventdna}, containing around 1,800 Dutch news articles. Despite their potential, both datasets have faced criticism regarding the reliability of manually annotated labels, as highlighted in related studies that employed these datasets for machine learning experiments \cite{de2020news,fatemi2023evaluating}. Beyond problems with reliability, these datasets exhibit additional limitations. The MN-DS dataset was derived from the NELA-GT-2018 dataset \cite{norregaard2019nela}, which was originally developed for misinformation research. As a result, it includes sources known to disseminate fake news, which may compromise the performance of models trained on it when applied to mainstream news texts. The EventDNA dataset, on the other hand, is limited by its annotations being performed at the lowest levels of the IPTC Media Topic taxonomy. Specific sublabels do not always align well with top-level labels, which makes the dataset less useful for automatic text classification on top-level labels. Additionally, while EventDNA is freely accessible, restrictions on third-party sharing do not allow experiments with closed-source large language models.

Prior research has investigated a range of machine learning models for the automatic classification of texts into IPTC news topics. The experiments encompass both traditional non-neural models, such as logistic regression, the Naive Bayes classifier, and support vector machines \cite{petukhova2023mn,fatemi2023evaluating}, as well as deep learning techniques, particularly BERT-like Transformer models \cite{petukhova2023mn,de2020news,pranjic2020evaluation}. Most of these studies have demonstrated that Transformer-based models consistently achieve state-of-the-art performance, indicating their potential in the domain of news topic classification \cite{petukhova2023mn,de2020news}.

Recently introduced Generative Pretrained Transformer (GPT) models, commonly known as large language models (LLMs), have demonstrated impressive performance across a range of text classification tasks, even when used in a zero-shot prompting fashion that does not require any training data \cite{kuzman2023automatic,ljubevsic2024dialect,huang2023chatgpt}. Furthermore, recent research in the field of natural language processing (NLP) has embraced ``LLMs as catalysts for redefining the landscape of data annotation in machine learning and NLP'' \cite{tan2024large}. Specifically, researchers are exploring the potential of GPT-based annotation of training data to replace time-consuming manual annotation, with the aim of using these annotations to fine-tune or improve the performance of other GPT or BERT-like models \cite{tan2024large,zeng2024learning}, \cite{sun2024principle,ding2023gpt}. Although LLMs have been employed to generate and annotate instruction-tuning datasets \cite{wang2024codeclm}, \cite{xu2023baize,wang2023self}, their use for the annotation of text classification data has been less explored. Meng et al. \cite{meng2022generating} contributed to this area by employing a GPT model to generate both texts and labels for training datasets through a zero-shot prompting approach, and by fine-tuning a smaller language model on the generated training dataset. In contrast to this approach, our study refrains from generating synthetic data to mitigate potential LLM biases. Instead, we use authentic news articles sourced from the web and employ a GPT model exclusively for classifying these texts into Media Topic labels. This GPT-based data annotation approach was also shown to be promising in the domain of sentiment analysis, where student models fine-tuned on datasets annotated by a GPT model exhibited comparable performance to models fine-tuned on datasets annotated by human annotators \cite{ding2023gpt}.

A recent investigation into the applicability of GPT models for IPTC news topic classification was conducted by Fatemi et al. \cite{fatemi2023evaluating} who evaluated a GPT model on the MN-DS dataset \cite{petukhova2023mn}. The applicability of GPT models for topic annotation was evaluated by training various machine learning models on the GPT-annotated dataset and comparing their performance against models trained on the same dataset, but manually annotated. The findings indicate the promising performance of the GPT model used as a data annotator. However, it is important to note that the performance of the model was not directly evaluated on a manually annotated test set. Consequently, the accuracy of the GPT model in this task remains undetermined.

Building on existing research, our study explores the use of GPT-based data annotation to fine-tune a BERT-like topic classifier. This research addresses multiple gaps in the literature. First, we assess the performance of both the teacher GPT model and fine-tuned student models on reliably manually annotated data. Second, we compare the annotation reliability of the GPT model with the reliability of human annotators. Additionally, our research involves machine learning experiments on multiple languages, which offer valuable insights into the models' multilingual and cross-lingual performance.

\section{LLM-Based Dataset Development}
\label{sec:training-data}

\Figure[t!](topskip=0pt, botskip=0pt, midskip=0pt)[width=\textwidth]{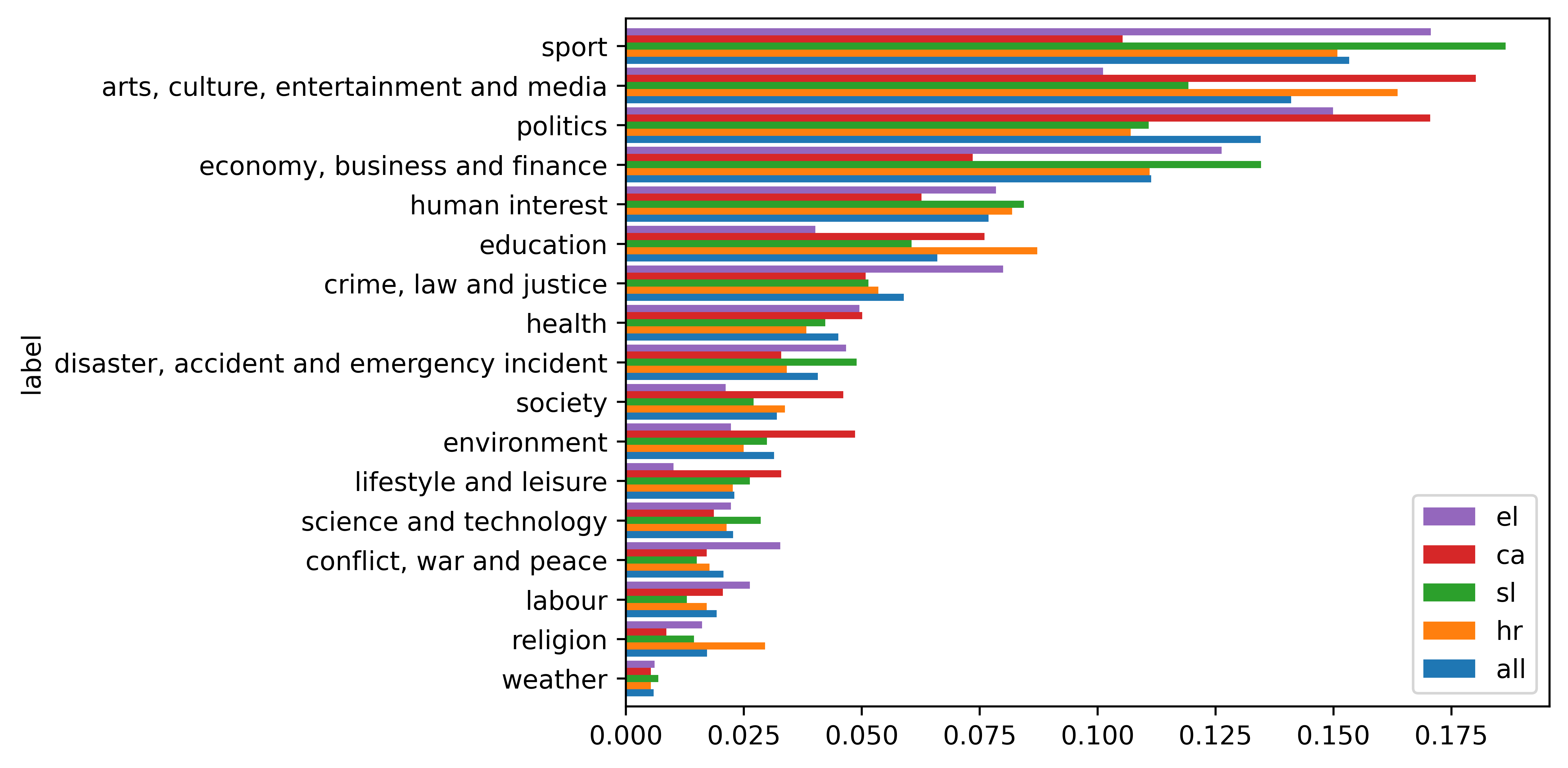}
{ \textbf{Distribution of labels in the GPT-annotated EMMediaTopic dataset comprising training and development splits (in percentages).}\label{fig:train-set-distribution}}

In this section, we present our methodology for the automatic annotation of training data that is used for fine-tuning a multilingual Transformer model which employs top-level IPTC Media Topic labels. The proposed methodology leverages three resources that have recently become available:
\begin{itemize}
    \item \textbf{MaCoCu Corpora} \cite{banon2022macocu}: The MaCoCu corpora collection comprises comparable web corpora in 10 less-resourced European languages, including Slovenian, Croatian, Catalan, Greek, Turkish, Icelandic, Ukrainian, and others. Developed through web crawling, MaCoCu corpora represent some of the largest text collections available for these languages. They encompass high-quality texts \cite{van2024language} from a wide range of sources and text genres \cite{kuzman2023get}. 
    \item \textbf{X-GENRE Classifier} \cite{kuzman2023automatic}: This multilingual Transformer-based genre classifier categorizes texts in various languages into a set of genre labels, such as \textit{News}, \textit{Promotion}, and \textit{Opinion/Argumentation}. The X-GENRE classifier allows us to efficiently extract relevant news content from general web corpora, which is crucial for the present task that is focused on news articles.
    \item \textbf{Multilingual GPT Models}: The multilingual Generative Pretrained Transformer (GPT) models have demonstrated impressive zero-shot classification capabilities across numerous languages and tasks. Inter alia, these models have shown high performance on South Slavic languages, including Croatian and Slovenian, which are included in our experiments \cite{ljubevsic2024jsi}. Specifically, we employ the GPT-4o model (version \texttt{gpt-4o-2024-05-13}) \cite{openai-gpt4o} provided by the OpenAI API, which outperformed other open- and closed-source GPT models in our preliminary experiments.
\end{itemize}

\subsection{Extraction of News Texts and Pre-Processing}

The pipeline for developing training and test data for news topic classification using the LLM teacher-student framework is illustrated in Fig. \ref{fig:dataset-pipeline}. The datasets used in this study are sourced from the Catalan (ca) \cite{macocu-ca}, Croatian (hr) \cite{macocu-hr}, Greek (el) \cite{macocu-el}, and Slovenian (sl) \cite{macocu-sl} MaCoCu web corpora that have been annotated with genres (genre-annotated datasets are available as part of the MaCoCu-Genre corpus collection \cite{macocu-genre}). Given the focus of IPTC topic classification on news articles, we extract from the web corpora the subset of news articles, identified with the X-GENRE classifier \cite{kuzman2023automatic}.
The X-GENRE classifier provides highly reliable classification into the \textit{News} genre, as well as eight other genre categories. Evaluation on a multilingual, manually annotated test set, sampled from the MaCoCu corpora, revealed F1 scores for the \textit{News} label ranging from 0.82 in Catalan to 0.95 in Croatian \cite{x-genre-classifier-huggingface}.
Furthermore, the dataset undergoes additional pre-processing, wherein the texts are truncated to the initial 512 words, which is a constraint imposed by the BERT-like models. 

\subsection{Automatic Annotation}

Automatic annotation involves using the GPT-4o model to assign 17 top-level IPTC Media Topic labels to news texts. We use labels from the Media Topic schema version from October 24, 2023. The annotation is applied to two separate samples, randomly extracted from the pre-processed news text collection described in the previous section: a first batch of 21,000 texts is annotated for the development of training and development datasets, and a second batch of 8,000 texts is annotated to create a test set. Both batches comprise equal numbers of instances in the four languages, namely Catalan (ca), Croatian (hr), Greek (el), and Slovenian (sl). 
The GPT-4o model is used in a zero-shot prompting manner. 
It is instructed to assign to each text one of the 17 main IPTC Media Topic labels, such as \textit{politics} and \textit{society}. The prompt (see Appendix \ref{sec:app-prompt}) was constructed based on the results of the preliminary experiments. It includes the task description and a list of labels with their descriptions (provided in Appendix \ref{sec:annotation-guidelines}). The annotation of 29,000 texts cost approximately 230€ and took approximately six  hours. This cost and time investment are deemed minimal compared to the resources that would be required for the manual annotation of a similar volume of instances.

\subsection{Training and Development Datasets}

To construct the training and development datasets, the first annotated batch of 21,000 instances is split into subsets of 20,000 and 1,000 texts, respectively, while maintaining stratification based on the GPT-assigned labels. The resulting dataset, which comprises both splits, has been published under the name ``the EMMediaTopic dataset'' in the CLARIN.SI repository (\url{http://hdl.handle.net/11356/1991}) \cite{emmediatopic}.

The distribution of labels within the EMMediaTopic dataset is shown in Fig. \ref{fig:train-set-distribution}. The dataset is shown to be relatively balanced by labels, with the most prevalent label (\textit{sport}) accounting for approximately 15\% of the instances. Moreover, the distribution of labels is consistent across the four languages. The least represented labels are \textit{weather}, accounting for less than 1\% of instances, and \textit{religion} and \textit{labour}, each assigned to less than 2\% of the texts.

\subsection{Test Dataset}

The test dataset is derived from the second batch of 8,000 automatically annotated texts, comprising texts in the same four languages as the training data, namely Slovenian, Croatian, Catalan, and Greek. To assess the models' performance across all 17 top IPTC Media Topic labels, the test set is balanced by GPT-assigned labels, which is achieved by selecting 18 instances per label per language (or fewer if there are fewer than 18 label instances in the batch). 
The selection of texts to be manually annotated comprised 1,199 instances with a balanced distribution across languages. Finally, the test set was manually annotated to obtain human gold labels. Details on the annotation campaign are provided in the next section.

\subsection{Teacher Model Performance}

The first application of the manually annotated test set was to measure the performance of the GPT-4o model that was used for automatic annotation. 
The model demonstrates consistently high performance, with an average macro-F1 score of 0.731 and a micro-F1 score of 0.722. Its performance across different languages is stable, with a 5-point difference between the highest macro-F1 score (on Slovenian) and the lowest (on Catalan), as shown in Table \ref{tab:GPT-performance}. 

\begin{table}
\caption{\textbf{GPT-4o performance on each language in the human-labeled test set in micro-F1 and macro-F1 scores, averaged across three prediction iterations, with standard deviation included.}}
\setlength{\tabcolsep}{3pt}
\begin{tabular}{|p{0.27\linewidth}|p{0.27\linewidth}|p{0.27\linewidth}|}
\hline
& {Micro-F1} & {Macro-F1} \\
\hline
Hr & 0.714 ± 0.002   &  0.721 ±    0.001        \\
Ca & 0.703 ± 0.002    &  0.702 ±    0.001          \\
Sl & 0.741 ± 0.000     &   0.748 ±    0.001            \\
El & 0.730 ± 0.009      &   0.738 ±   0.009      \\
\hline
Entire Test Set & 0.722 ± 0.002 & 0.731 ± 0.003 \\
\hline
\end{tabular}
\label{tab:GPT-performance}
\end{table}

\section{Manual Annotation of Test Data}
\label{sec:test-data}

The test sample, balanced by GPT-assigned labels and the four languages, 
 is manually annotated by a single annotator. The annotator was provided with annotation guidelines with the descriptions of the labels (see Appendix \ref{sec:annotation-guidelines}) and received brief training on the annotation process. The description of the labels is based on the official definitions provided by the IPTC \cite{iptc-topic-definitions}. However, since these definitions offer only a high-level understanding of the distinctions between labels, we enriched the descriptions with examples of their lower-level labels to facilitate a more precise and consistent annotation.

In addition to the standard 17 top-level IPTC Media Topic labels, three additional labels have been introduced to assist the annotator in identifying and excluding texts that are not suitable for the purposes of our task: \textit{do not know} (highly challenging instance lacking a clear topic), \textit{not news} (text of a different genre, e.g., \textit{Promotion}), and \textit{multiple} (presence of multiple texts in a single instance). The manual annotation process excluded 70 instances (5.83\%) as unsuitable. Among these, 4\% were excluded because of the absence of a clear topic, 2\% because they contained multiple texts in a single instance, and 0.1\% due to incorrect genre classification. The final test sample consists of 1,129 text instances. The test dataset is available on request from the corresponding author. It is not publicly released to prevent its integration into large language models (LLMs) during their pretraining or fine-tuning phase to maintain the integrity of future model evaluations.

\subsection{Inter-Annotator Agreement}
\label{sec:inter-annotator}

To assess the reliability of human and LLM annotations, a subset of 339 instances from the test set, balanced by labels and languages, was annotated by a second annotator.
Human annotators received the same guidelines, provided in Appendix  \ref{sec:annotation-guidelines}, and participated in a preliminary test annotation which was followed by a discussion to resolve the disagreements.

The inter-annotator agreement is evaluated using Krippendorff's alpha metric~\cite{krippendorff2018content}, a statistical measure frequently employed in content analysis, social sciences, and machine learning to determine the reliability of multiple human annotators in data classification. The metric accounts for the likelihood of agreement occurring by chance, with a value of 1 indicating perfect agreement and a value of 0 indicating agreement equal to chance \cite{krippendorff2004measuring}. Theoretically, reliable inter-annotator agreement is represented by alpha values of 0.8 or higher, whereas an alpha of 0.667 is considered the threshold for acceptable annotation reliability~\cite{krippendorff2018content}. Nonetheless, this threshold value is frequently not achieved in manual annotation campaigns for text classification tasks. This shortcoming can be attributed to the high subjectivity and complexity of annotation tasks, which often involve cases where characteristics of multiple labels are present \cite{klie2024analyzing}. This problem has also been demonstrated in tasks similar to topic classification, such as automatic genre identification \cite{kuzman2023survey}, sentiment annotation \cite{van2021validity}, \cite{lind2017content,pelicon2020zero}, and detection of hate speech \cite{leite2020toxic,ljubevsic2019frenk}.

In our experiments, the nominal Krippendorff's alpha between the two annotators reached 0.728. This outcome is deemed satisfactory as it exceeds the threshold for acceptable reliability. Furthermore, this level of inter-annotator agreement is comparable to or exceeds the agreement observed in similar text-level classification tasks. At the same time, the moderate level of agreement highlights the challenging nature of this task, as multiple topic labels may be discussed in certain news articles.


As shown in Table \ref{tab:inter-annotator-agreement}, the inter-annotator agreement between each of the human annotators and the LLM model was on par with the agreement between the two human annotators. This finding addresses Research Question 1 (RQ1), demonstrating that the LLM model is as capable of data annotation for this task as human annotators. 

\begin{table}
\caption{\textbf{Pair-wise inter-annotator agreement in terms of the nominal Krippendorff's alpha.}}
\setlength{\tabcolsep}{3pt}
\begin{tabular}{|p{0.45\linewidth}|p{0.45\linewidth}|}
\hline
Annotators                          & Krippendorff's alpha \\
\hline
1st ann \& 2nd ann  & 0.728                            \\
1st ann \& GPT-4o      & 0.693 \\
2nd ann \& GPT-4o & 0.752                           \\
\hline
\end{tabular}
\label{tab:inter-annotator-agreement}
\end{table}

\Figure[t!](topskip=0pt, botskip=0pt, midskip=0pt)[width=0.45\textwidth]{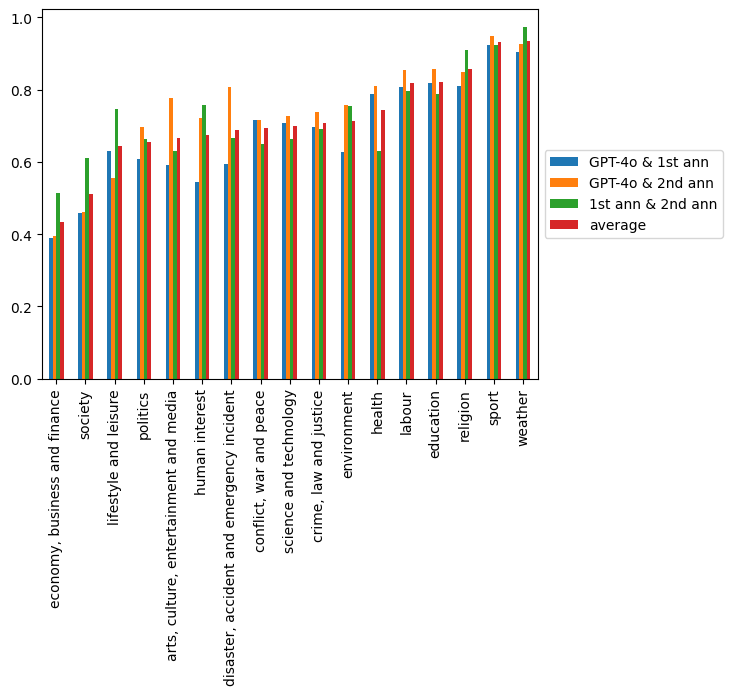}
{ \textbf{Label-level inter-annotator agreement in terms of the nominal Krippendorff's alpha.}\label{fig:label-level-agreement}}

We also assess the agreement between the human and LLM annotators at the label level. Fig. \ref{fig:label-level-agreement} illustrates the substantial variation in inter-annotator agreement across topic labels, highlighting labels that are less clearly defined, which leads to confusion among annotators. The labels \textit{sport} and \textit{weather} are well comprehended by both human and LLM annotators, achieving high inter-annotator agreement above 0.90 in terms of Krippendorff's alpha. Labels \textit{religion}, \textit{lifestyle and leisure}, and \textit{society} present challenges for the LLM, with human annotators demonstrating significantly higher agreement. The label \textit{economy, business and finance} is particularly problematic for both humans and the LLM, with an average inter-annotator agreement of 0.43.

\begin{table}
\caption{\textbf{Intra-annotator agreement in terms of the nominal Krippendorff's alpha for the human annotator (1st annotator) and the LLM teacher model.}}
\setlength{\tabcolsep}{3pt}
\begin{tabular}{|p{0.45\linewidth}|p{0.45\linewidth}|}
\hline
Annotator                          & Krippendorff's alpha \\
\hline
GPT-4o          & 0.934                      \\
Human annotator (1st ann) & 0.796     \\            
\hline
\end{tabular}
\label{tab:intra-annotator-agreement}
\end{table}

\subsection{Intra-Annotator Agreement}

To further assess the reliability of the annotations provided by humans and LLMs, we instruct both a human annotator and the GPT-4o model to re-annotate the subset that was previously used to calculate the inter-annotator agreement. Annotators are given the same instructions as in the initial annotation process, and the labels previously assigned to the texts are hidden from them. Intra-annotator agreement is evaluated using Krippendorff's alpha measuring consistency between two annotation runs conducted by the same annotator. Table \ref{tab:intra-annotator-agreement} presents the intra-annotator agreement for the human and LLM annotators. The human annotator achieved a relatively high Krippendorff's alpha of approximately 0.80, indicating a reliable level of self-agreement. 
The GPT-4o model demonstrated even greater consistency with Krippendorff's alpha of 0.93, highlighting the reliability of large language models in producing consistent responses. However, while such consistency is advantageous, more diverse responses, similar to those obtained from the human annotator, can be beneficial for various research directions, including learning from disagreement~\cite{uma2021learning}.

\section{Student Model Fine-Tuning Experiments}
\label{sec:experiments}

In this section, we describe the experiments that involve fine-tuning of student BERT-like models on training data that were automatically annotated by the teacher GPT model. In Section \ref{sec:model}, we present the student model which is used in the experiments that examine (1) the impact of training data size on classification performance (Section \ref{sec:exp-data-size}), (2) the model's label-level performance (Section \ref{sec:label-level}), and (3) the model's monolingual, multilingual, and cross-lingual capabilities (Section \ref{sec:exp-monolingual}).

\subsection{Student Model}
\label{sec:model}

As the student model, we select the large-sized XLM-RoBERTa model \cite{conneau2020unsupervised} (available on the Hugging Face repository: \url{https://huggingface.co/FacebookAI/xlm-roberta-large}), which we fine-tune on the teacher-annotated EMMediaTopic dataset. This BERT-like encoder model, pretrained on hundred languages, has demonstrated strong multilingual and cross-lingual performance across various text classification tasks, which also involved the languages used in our experiments \cite{ulvcar2021evaluation,van2024language,kuzman2023automatic}. The optimal hyperparameters were identified via a hyperparameter search performed on the development dataset. Appendix \ref{sec:app-hyperparameters} provides a comprehensive description of the selected hyperparameters.

The models, fine-tuned on the teacher-annotated training dataset, are evaluated on the test dataset via micro-F1 and macro-F1 scores to assess the performance at both the instance and label levels, respectively. Each model was trained and tested at least three times to gain an understanding of the performance consistency. The reported scores represent the mean values calculated across all iterations and are accompanied by a standard deviation. In the experiments related to the training data size, five iterations were performed per data point owing to the addition of the subset selection process, which was an additional source of variation of the results.


\subsection{Impact of the Training Data Size}
\label{sec:exp-data-size}

In this section, we address Research Question 2 (RQ2). First, we explore the feasibility of achieving high performance of BERT-like models on the news topic classification task when fine-tuned on automatically annotated data. Second, we investigate the minimum amount of training data required to attain a performance level comparable to that of the teacher model. To this end, we fine-tune the XLM-RoBERTa model on the subsets from the original training set, which are stratified by label size and balanced across languages. The smaller training datasets comprise 1,000, 2,500, 5,000, 10,000, and 15,000 instances. 

\Figure[t!](topskip=0pt, botskip=0pt, midskip=0pt)[width=0.4\textwidth]{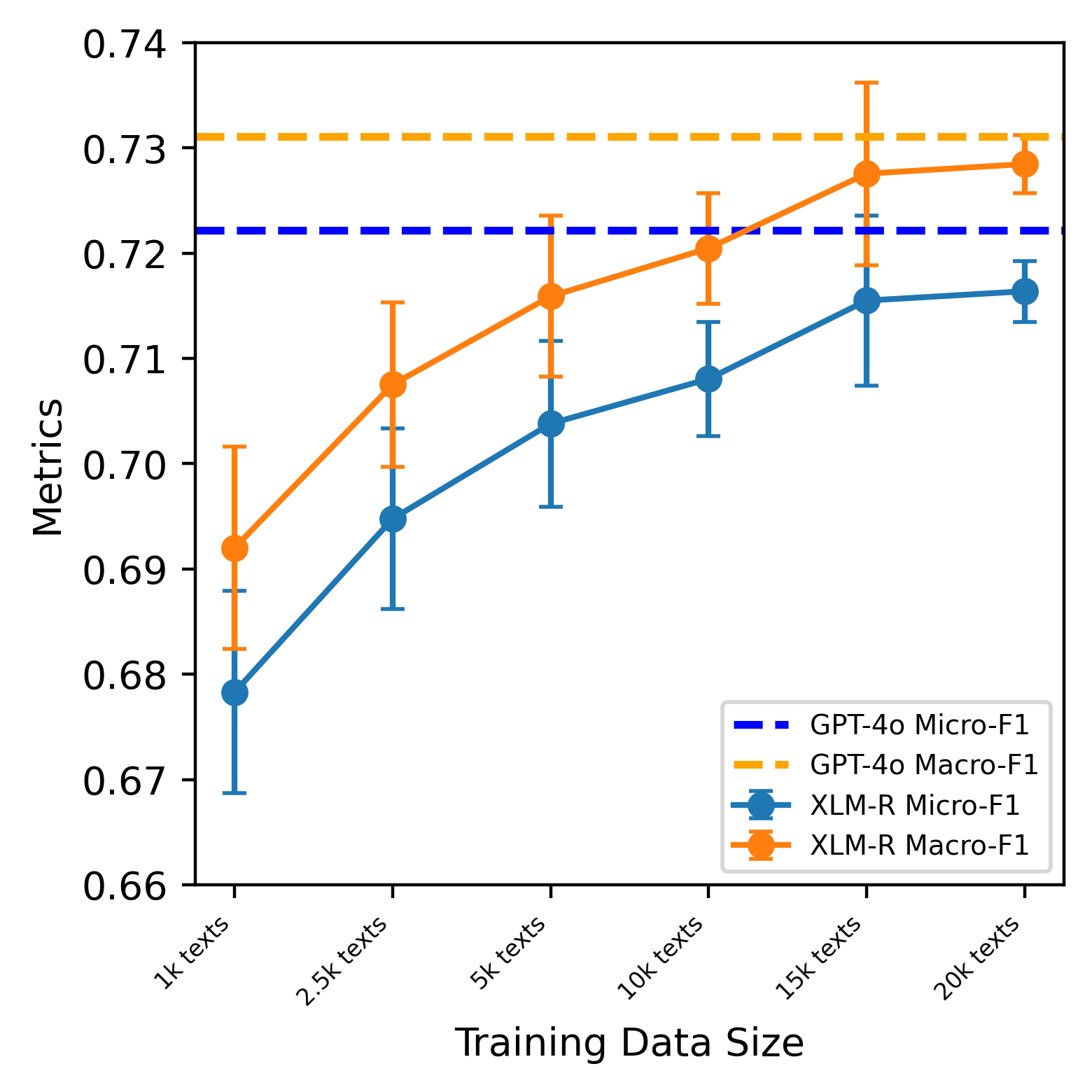}
{ \textbf{Performance in micro-F1 and macro-F1 scores of the XLM-RoBERTa (XLM-R) model fine-tuned on various sizes of training data, compared to the zero-shot GPT-4o performance as the upper limit. The scores are averaged across five iterations of fine-tuning and evaluation, each using different random sample of a specified size, drawn from the training dataset.}\label{fig:model-size-comparison}}

Fig. \ref{fig:model-size-comparison} shows the impact of the training data size on the model performance. The performance of the GPT-4o model is considered to be an upper limit, as it served as the teacher model for annotating the training dataset for the student model. The level of improvement begins to plateau after 15,000 instances. Notably, student models trained on datasets comprising 15,000 data points or more exhibit performance levels that are comparable to those of the teacher GPT model. There is less than a one-point difference in macro-F1 and micro-F1 scores between the student and teacher models. These findings confirm the feasibility of achieving high performance in the news topic classification task using a student model trained on data annotated by a teacher model.

Fine-tuning student models on training datasets comprising 15,000 data points or more provides optimal results; however, the findings also reveal that the models exhibit substantial performance even with minimal training data. Specifically, student models trained on datasets comprising 1,000--5,000 instances achieve average micro-F1 scores ranging from 0.678 to 0.704 and macro-F1 scores between 0.692 and 0.716. 
Notably, the student model trained on only 1,000 instances demonstrates a performance that is only four points lower than that of the teacher GPT model.

\Figure[t!](topskip=0pt, botskip=0pt, midskip=0pt)[width=0.48\textwidth]{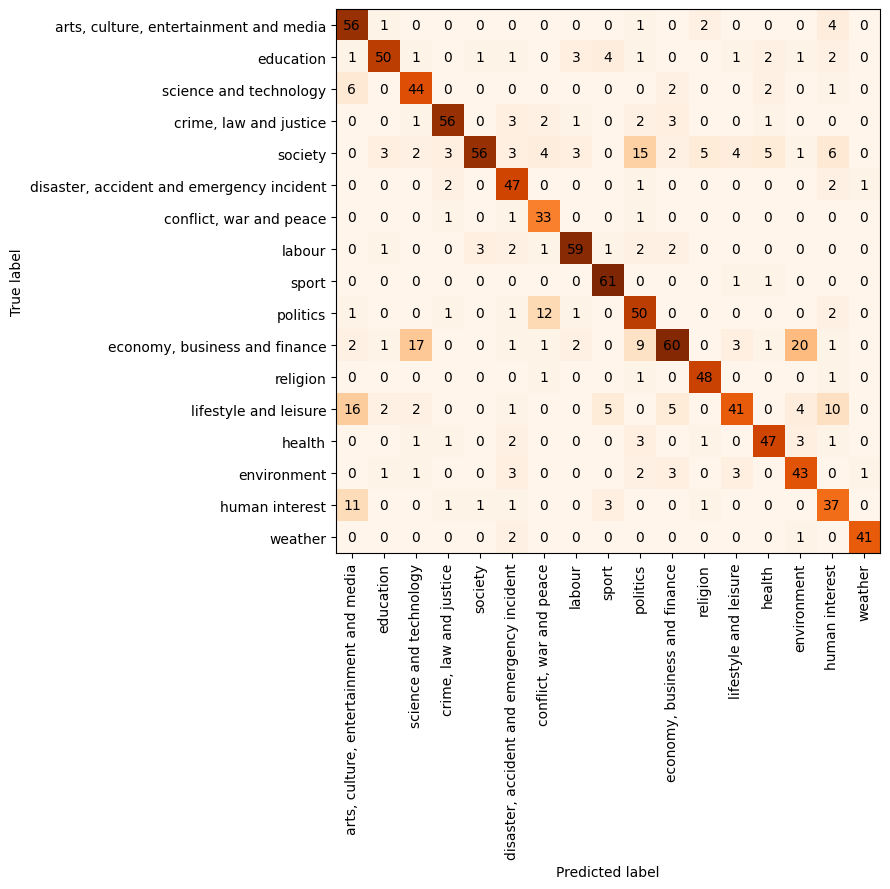}
{ \textbf{Confusion matrix for the best-performing XLM-RoBERTa student model.}\label{fig:confusion-matrix}}

\begin{table*}
\caption{\textbf{Performance in macro-F1 scores of monolingual and multilingual models trained on 5,000 instances and evaluated on each language in the test split, and on the entire test split. The results of the monolingual training and evaluation are highlighted in gray color. The highest score for each language-specific test set is highlighted in bold. GPT-4o performance is added as the upper limit, as the models were trained on its predictions. The reported scores represent the average results across three iterations of training and evaluation, with the standard deviation provided.}}
\setlength{\tabcolsep}{3pt}
\begin{tabular}{|p{0.15\linewidth}|p{0.15\linewidth}|p{0.15\linewidth}|p{0.15\linewidth}|p{0.15\linewidth}|p{0.15\linewidth}|}
\hline
& Hr Test   & Ca Test   & Sl Test  & El Test & Entire Test Set   \\
\hline
Hr Model & \cellcolor[gray]{0.8}0.701 ±   0.017  & \textbf{0.672} ±    0.005 & 0.733 ±   0.014  & \textbf{0.739} ±    0.011 & 0.716  ±  0.009\\
Ca Model & 0.706 ±    0.012 & \cellcolor[gray]{0.8}0.671 ±    0.008  & 0.728 ±    0.005 & 0.715  ±    0.014 & 0.710  ±   0.007 \\
Sl Model & \textbf{0.711}  ±   0.007  & 0.660 ±    0.016 & \cellcolor[gray]{0.8}0.736 ±    0.014 & 0.721 ±  0.013  & 0.713  ±  0.005 \\
El Model & 0.674  ±    0.004 & 0.662  ±    0.012 & 0.716 ±    0.011 & \cellcolor[gray]{0.8}0.706 ±    0.013 & 0.695 ± 0.006\\
Multilingual    & 0.707 ±    0.011 & 0.656 ±    0.024 & \textbf{0.741} ±    0.007 & 0.729 ±    0.004 & 0.714 ±    0.009\\
\hline
GPT-4o       & 0.721 ±    0.001 & 0.702 ±    0.001 & 0.748 ±    0.001 & 0.738 ±   0.009 & 0.731 ±   0.003\\
\hline
\end{tabular}
\label{tab:monolingual-experiments}
\end{table*}

\subsection{Label-Level Performance}
\label{sec:label-level}
The best-performing version of the XLM-RoBERTa student model, namely, a version trained on 15,000 instances, has been made available in the Hugging Face repository (accessible at \url{https://huggingface.co/classla/multilingual-IPTC-news-topic-classifier}). This model attains a macro-F1 score of 0.746 and micro-F1 score of 0.734. Further insights into the performance of the model at the label level can be obtained from the confusion matrix depicted in Fig. \ref{fig:confusion-matrix}, which displays the true versus predicted labels.

The labels that are most accurately predicted by the model are \textit{weather}, \textit{sport}, and \textit{religion}, with label-level F1 scores above 0.89. These findings are consistent with the label-level inter-annotator agreement shown in Fig. \ref{fig:label-level-agreement} (see Section \ref{sec:inter-annotator}), which identified these labels as those with the highest consensus among human annotators. Interestingly, despite the \textit{weather} label being represented by less than 1\% of the instances of the training dataset, this did not negatively impact the performance of the model. The model demonstrated a near-perfect classification accuracy for this label, achieving an F1 score of 0.94.
This result confirms the usefulness of the classifier also for the labels that are less represented in the training dataset.

In contrast, the most challenging labels for the XLM-RoBERTa student model are \textit{lifestyle and leisure}, \textit{human interest}, and \textit{economy, business and finance}, as evidenced by label-level F1 scores ranging from 0.59 to 0.62. This is not surprising, given that these categories have also been identified as challenging for both human and LLM annotators. As shown in Fig. \ref{fig:confusion-matrix}, the label \textit{economy, business and finance} is frequently misclassified as \textit{science and technology} when the texts pertain to the sales of technology products; or as \textit{environment} in instances where the texts address electricity infrastructure. The label \textit{human interest} is often mistaken for \textit{arts, culture, entertainment and media}, because both categories encompass content related to celebrities. The former is typically associated with tabloid-style presentations of celebrity lives, whereas the latter is intended to cover more serious topics. Similarly, the label \textit{lifestyle and leisure} frequently overlaps with both \textit{human interest} and \textit{arts, culture, entertainment and media}, as it encompasses individuals' recreational activities, which may intersect with topics represented by the other two labels. These overlaps illustrate the complexities inherent in IPTC topic classification, which pose challenges to both manual annotation and automatic classification. These challenges arise primarily from the broad nature of top-level labels, which often leads to intersections among categories. A potential solution to these problems is to conduct classification with labels from the lower levels of the Media Topic schema, which are more specific but also more challenging due to their high number.

\subsection{Monolingual, Multilingual and Cross-Lingual Performance}
\label{sec:exp-monolingual}

The preceding section demonstrated the potential for achieving high performance using the LLM teacher-student setup for the task of news topic classification. This section explores the applicability of student models to languages on which they were not fine-tuned. We analyze the performance of fine-tuned models in monolingual, multilingual, and cross-lingual settings, addressing Research Questions 3 (RQ3) and 4 (RQ4).

Specifically, we fine-tune the XLM-RoBERTa model on monolingual subsets extracted from the original training dataset comprising only Croatian (hr), Slovenian (sl), Catalan (ca), or Greek (el) instances. Each monolingual subset consists of 5,000 instances in the target language, and the subsets have similar label distributions. Additionally, the evaluation includes the multilingual model, trained on samples comprising 5,000 instances (as described in Section \ref{sec:exp-data-size}), distributed equally across the four languages. The training dataset for the multilingual model is limited to 5,000 instances to ensure that the multilingual model is trained on a dataset of equivalent size to that of the monolingual models and to eliminate any potential influence of the dataset size on the results.

The models are evaluated on the manually annotated test dataset, split into target language subsets. Each monolingual test subset comprises approximately 300 instances. The reported scores are averaged across the three iterations of fine-tuning and evaluation.

Table \ref{tab:monolingual-experiments} presents the performance of the monolingual and multilingual models across four languages included in the test set. In a zero-shot cross-lingual scenario -- where the models are evaluated on a language different from the one they were fine-tuned on --, the models demonstrate relatively high macro-F1 scores, ranging from 0.66 to 0.74. Moreover, the zero-shot cross-lingual performance is often comparable to, and in most cases even exceeds, the models' performance in monolingual and multilingual settings -- where the models are evaluated on the language which is included in their (either monolingual or multilingual) fine-tuning dataset. These findings, pertaining to Research Question 3 (RQ3), demonstrate that student models exhibit high performance even when applied to languages that are not part of the fine-tuning training dataset.


Second, we compare the performance of student models on a target language when they are fine-tuned on 5,000 instances of (1) monolingual data in the target language, or (2) multilingual data that are balanced across the target language and three additional languages. Through these experiments, we address Research Question 4 (RQ4), which focuses on investigating whether the development of language-specific models offers any advantages over the development of a multilingual model trained on an equivalent amount of data, consisting of instances in the target language and other languages. If the experiments would show that monolingual results are significantly higher, this might require to host potentially many language-specific models for cases where top performance is important enough. The results presented in Table \ref{tab:monolingual-experiments}, where the monolingual performance is highlighted in gray, however, reveal that the multilingual model achieves performance that is comparable to, or on certain language-specific test sets surpasses that of the model fine-tuned only on the target language. This finding highlights the benefits of training student models on multiple languages, through which the robustness of the model is quite likely to improve. More importantly, this result allows the final inference procedure to be very simple, with a single multilingual model that can be applied to any text written in one of the 100 supported languages on which the XLM-RoBERTa model was pretrained \cite{conneau2020unsupervised}. 

\section{Conclusion}
\label{sec:conclusions}

This paper presents a novel methodology for developing annotated training data using a teacher-student framework based on large language models. By fine-tuning a smaller Transformer model on the data annotated by a GPT model, this methodology allows for scalable classification of news topics without the need for manual annotation of the training data. The code for the development of GPT-annotated training data and XLM-RoBERTa topic classifiers is publicly accessible (\url{https://github.com/TajaKuzman/IPTC-Media-Topic-Classification}).

The IPTC Media Topic training dataset is developed by extracting news articles from web corpora in four languages (Slovenian, Croatian, Greek, and Catalan) based on automatic genre identification using the X-GENRE classifier \cite{kuzman2023automatic} and automatic topic annotation using the GPT model as a teacher model. The EMMediaTopic dataset \cite{emmediatopic} has been made freely available in the CLARIN.SI repository (\url{http://hdl.handle.net/11356/1991}).

The GPT model, used in a zero-shot prompting fashion, demonstrates high performance across all languages and labels, with an average macro-F1 score of 0.731 and micro-F1 score of 0.722. The inter-annotator agreement between the two human annotators is comparable to the agreement between the teacher LLM model and each of the human annotators. Furthermore, the intra-annotator agreement analysis reveals that the GPT model demonstrates a significantly higher consistency in topic labeling than the human annotator. The results indicate that large language models possess a comparable ability to human annotators in labeling training data with news topic categories, but with less variation in the output.

GPT models are computationally too demanding for scaling topic annotation to millions of data points that regularly need to be annotated, as is the case in the news media industry. Thus, we develop smaller BERT-like models, considered as student models, by fine-tuning them on the GPT-annotated EMMediaTopic training dataset. We show that the developed training data are of a sufficient quality to achieve high performance of the student models. This finding confirms the feasibility of the LLM teacher-student framework for the development of multilingual news topic classifiers. Specifically, the XLM-RoBERTa-based student models, when fine-tuned on 15,000 or more data points, demonstrate a performance comparable to that of the GPT teacher model, with a difference of less than one point in both the micro-F1 and macro-F1 scores. Motivated by positive results, we publish the best-performing student model on Hugging Face (\url{https://huggingface.co/classla/multilingual-IPTC-news-topic-classifier}), providing the first open-source IPTC Media Topic classification model for the top layer of the Media Topic schema that can be applied to any language covered by the XLM-RoBERTa model. The model was trained on 15,000 instances in four languages, and it achieves a micro-F1 score of 0.734 and a macro-F1 score of 0.746.

Furthermore, we delve deeper into the constraints of the performance of the student models, investigating their ability to generalize in a zero-shot cross-lingual scenario. The experiments with student models fine-tuned either on monolingual or multilingual training data and evaluated on four languages show high zero-shot cross-lingual performance, yielding macro-F1 scores between 0.66 to 0.74. Moreover, the zero-shot cross-lingual performance exceeds the models' performance in monolingual and multilingual settings. These results indicate that incorporating the target language within the training dataset is not essential for achieving a high topic classification performance in that language.

Further experiments concerning the scalability of IPTC classification across multiple languages investigate the potential advantages of developing language-specific student models compared with constructing a multilingual model fine-tuned on the same amount of data in four languages, thereby less data in the target language. An analysis of the performance of monolingual versus multilingual models across four target languages reveals that the multilingual model achieves a performance that is either comparable or superior to that of models fine-tuned exclusively on the target language. This finding suggests that multilingual models are a more effective approach to catering to multiple languages. These models provide high performance across multiple languages while requiring less annotated data in the target language and incurring smaller processing and storage costs compared to the development of individual models for each language.

In this study, we employ only the 17 top-level categories of the IPTC Media Topic hierarchical schema. The analysis of the agreement between human annotators and the LLM annotator for each category, as well as the confusion matrix for the fine-tuned model, highlighted challenges associated with the ambiguity of certain top-level labels. In future work, we plan to extend our experiments to include more fine-grained categories from the lower levels of the IPTC schema, which are more useful for some users and news media providers. We consider following the hierarchical approach proposed by \cite{fatemi2023evaluating} and \cite{de2020news}, where the top label is assigned to the text first with the classifier presented in this paper, and then based on the predicted label, a more specialized lower-layer classifier is selected to make deeper level-two predictions. 
Moreover, our current experiments involve single-label classification, which does not account for the multifaceted nature of some texts in which multiple top-level topics could be useful to the end user. To address this, we plan to move towards a multi-label classification setting based on the information on the confidence of the student model.

Motivated by positive results, we intend to expand our experiments involving the LLM teacher-student framework to topic classification across various domains, such as parliamentary proceedings and web corpora. Furthermore, we aim to explore the applicability of the proposed methodology to additional text classification tasks, including automatic genre identification, which is recognized as a challenging task for human annotators \cite{kuzman2023survey}. 

Furthermore, future work could include a more in-depth exploration of potential biases that might emerge in the fine-tuned student model. The biases may be introduced from the original sources, since the training data is sourced from a large collection of online news articles, as well as from the large language model used for annotating these training data. Given that the news topic categories include sensitive subjects such as \textit{conflict, war and peace}, \textit{religion}, and \textit{society}, it is important to consider the potential biases originating from both the training data and the large language model while deploying the automatic news classification system in real-world applications.

\section{Limitations}

While our LLM teacher-student methodology provides promising results, it is important to acknowledge that the success of this approach is fundamentally dependent on the credibility of the teacher model, as its performance represents the upper limit of what can be achieved with the framework. In our study, the GPT model demonstrated strong performance in the news topic classification task, achieving high macro-F1 and micro-F1 scores and exhibiting annotation reliability comparable to that of human annotators. These findings highlight the feasibility of the framework for this specific task, providing high-quality annotations at minimal cost. However, this level of performance may not necessarily be replicable across other natural language processing tasks or languages, where the conventional method of training on manually annotated data may still be preferable. Therefore, it is crucial that researchers assess the teacher model on manually annotated test datasets to ensure its reliability and comparability to human annotators prior to applying the LLM teacher-student approach to their use cases.

\clearpage
\appendices

\section{\break Annotation Guidelines}
\label{sec:annotation-guidelines}

\paragraph{General Instructions} Annotate the provided texts with the IPTC topic labels, provided below. If it is hard to choose between two labels, you can also help yourself by searching through the taxonomy of sublabels (\url{https://www.iptc.org/std/NewsCodes/treeview/mediatopic/mediatopic-en-GB.html}) where you might find the specific topic of your text and see under which of out top-level labels it fits (but do not spend too much time on it, if it is very hard to decide, rather annotate it as ``do not know'').

\paragraph{Additional Labels for Hard Cases} We are interested only in reasonable and informative instances. That is why we added to the list of labels three more categories for you to ``discard'' the text:
\begin{itemize}
    \item \textit{do not know}: if it is impossible to decide under which label this instance fits.
    \item \textit{not news}: if the text is not a news article, for example, it is an advertisement, recipe, legal text, blog post, etc.
    \item \textit{multiple}: if the instance is not a single text, but consists of multiple texts.
\end{itemize}

\subsection*{IPTC Label Description}
\label{sec:label-description}

Table \ref{tab:IPTC-schema} presents the IPTC Media Topic labels along with their descriptions, as provided to the human annotators and GPT model.

\begin{table*}
\caption{\textbf{IPTC Media Topic labels and their descriptions, which have been constructed by the authors of this paper.}}
\setlength{\tabcolsep}{3pt}
\begin{tabular}{|p{0.2\linewidth}|p{0.8\linewidth}|}
\hline
Label                                    & Description  \\
\hline
arts, culture, entertainment and media    & News about cinema, dance, fashion, hairstyle, jewellery, festivals, literature, music, theatre, TV shows, painting, photography, woodworking, art exhibitions, libraries and museums, language, cultural heritage, news media, radio and television, social media, influencers, and disinformation.                                                                                                                                                                                                                                                            \\\hline
conflict, war and peace                   & News about terrorism, wars, wars victims, cyber warfare, civil unrest (demonstrations, riots, rebellions), peace talks and other peace activities.                                                                                                                                                                                                                                                                                                                                                                                                             \\\hline
crime, law and justice                    & News about committed crime and illegal activities, the system of courts, law and law enforcement (e.g., judges, lawyers, trials, punishments of offenders).                                                                                                                                                                                                                                                                                                                                                                                                    \\\hline
disaster, accident and emergency incident & Man-made or natural events resulting in injuries, death or damage, e.g., explosions, transport accidents, famine, drowning, natural disasters, emergency planning and response.                                                                                                                                                                                                                                                                                                                                                                                \\\hline
economy, business and finance             & News about companies, products and services, any kind of industries, national economy, international trading, banks, (crypto)currency, business and trade societies, economic trends and indicators (inflation, employment statistics, GDP, mortgages, ...), international economic institutions, utilities (electricity, heating, waste management, water supply).                                                                                                                                                                                            \\\hline
education                                 & All aspects of furthering knowledge, formally or informally, including news about schools, curricula, grading, remote learning, teachers and students.                                                                                                                                                                                                                                                                                                                                                                                                         \\\hline
environment                               & News about climate change, energy saving, sustainability, pollution, population growth, natural resources, forests, mountains, bodies of water, ecosystem, animals, flowers and plants.                                                                                                                                                                                                                                                                                                                                                                        \\\hline
health                                    & News about diseases, injuries, mental health problems, health treatments, diets, vaccines, drugs, government health care, hospitals, medical staff, health insurance.                                                                                                                                                                                                                                                                                                                                                                                          \\\hline
human interest                            & News about life and behaviour of royalty and celebrities, news about obtaining awards, ceremonies (graduation, wedding, funeral, celebration of launching something), birthdays and anniversaries, and news about silly or stupid human errors.                                                                                                                                                                                                                                                                                                                 \\\hline
labour                                    & News about employment, employment legislation, employees and employers, commuting, parental leave, volunteering, wages, social security, labour market, retirement, unemployment, unions.                                                                                                                                                                                                                                                                                                                                                                      \\\hline
lifestyle and leisure                     & News about hobbies, clubs and societies, games, lottery, enthusiasm about food or drinks, car/motorcycle lovers, public holidays, leisure venues (amusement parks, cafes, bars, restaurants, etc.), exercise and fitness, outdoor recreational activities (e.g., fishing, hunting), travel and tourism, mental well-being, parties, maintaining and decorating house and garden.                                                                                                                                                                               \\\hline
politics                                  & News about local, regional, national and international exercise of power, including news about election, fundamental rights, government, non-governmental organisations, political crises, non-violent international relations, public employees, government policies.                                                                                                                                                                                                                                                                                         \\\hline
religion                                  & News about religions, cults, religious conflicts, relations between religion and government, churches, religious holidays and festivals, religious leaders and rituals, and religious texts.                                                                                                                                                                                                                                                                                                                                                                   \\\hline
science and technology                    & News about natural sciences and social sciences, mathematics, technology and engineering, scientific institutions, scientific research, scientific publications and innovation.                                                                                                                                                                                                                                                                                                                                                                                \\\hline
society                                   & News about social interactions (e.g., networking), demographic analyses, population census, discrimination, efforts for inclusion and equity, emigration and immigration, communities of people and minorities (LGBTQ, older people, children, indigenous people, etc.), homelessness, poverty, societal problems (addictions, bullying), ethical issues (suicide, euthanasia, sexual behaviour) and social services and charity, relationships (dating, divorce, marriage), family (family planning, adoption, abortion, contraception, pregnancy, parenting). \\\hline
sport                                     & News about sports that can be executed in competitions, e.g., basketball, football, swimming, athletics, chess, dog racing, diving, golf, gymnastics, martial arts, climbing, etc.; sport achievements, sport events, sport organisation, sport venues (stadiums, gymnasiums, ...), referees, coaches, sport clubs, drug use in sport.     \\ \hline 
weather                                   & News about weather forecasts, weather phenomena and weather warning.                                                                       \\
\hline
\end{tabular}
\label{tab:IPTC-schema}
\end{table*}


\section{\break Prompt for automatic annotation}
\label{sec:app-prompt}

Fig. \ref{fig:prompt} shows the prompt presented to the GPT model for the purpose of automatic annotation. The prompt includes descriptions of the labels, presented in Table \ref{tab:IPTC-schema}.

\Figure[t!](topskip=0pt, botskip=0pt, midskip=0pt)[width=0.9\textwidth]{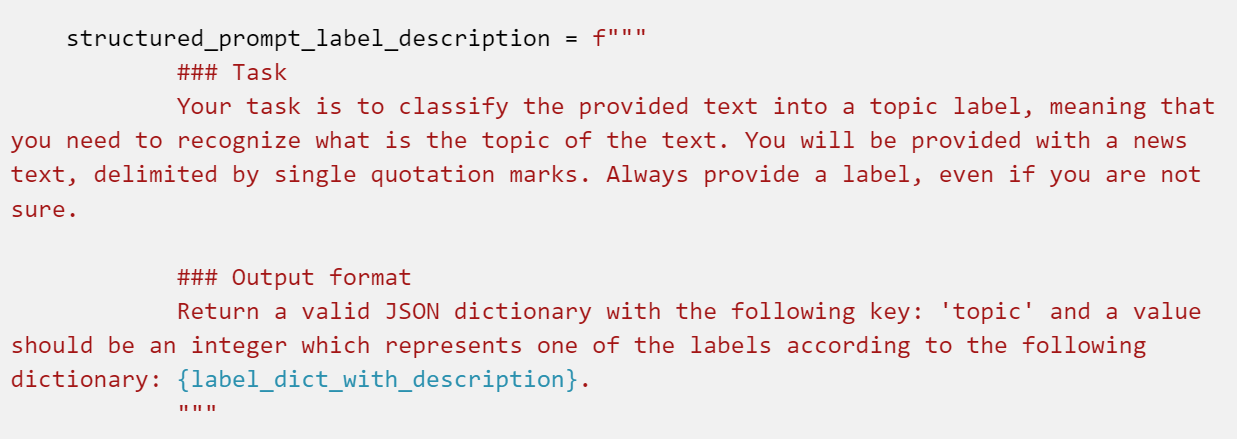}
{ \textbf{Prompt used for automatic annotation of data with the GPT-4o model.}\label{fig:prompt}}

\clearpage

\section{\break Fine-tuning hyperparameters for the XLM-RoBERTa models}
\label{sec:app-hyperparameters}

\begin{table}[!htpb]
\caption{\textbf{Number of epochs used for fine-tuning the XLM-RoBERTa model, depending on the size of the training data (number of instances).}}
\setlength{\tabcolsep}{3pt}
\begin{tabular}{|p{0.45\linewidth}|p{0.45\linewidth}|}
\hline
Training Data Size  & Epoch Number \\
\hline
20,000     & 3            \\
15,000     & 5            \\
10,000     & 9            \\
5,000      & 10           \\
2,500      & 22           \\
1,000      & 24   \\                        
\hline
\end{tabular}
\label{tab:epoch-numbers}
\end{table}

In the fine-tuning experiments, we use the following hyperparameters that were determined through a hyperparameter search on the development dataset: a learning rate of $8e^{-6}$, a train batch size of 32, and a maximum sequence length of 512 tokens. The optimal number of training epochs was found to depend on the size of the training data, as shown in Table \ref{tab:epoch-numbers}.

\clearpage


\bibliographystyle{IEEEtran}
\bibliography{bibliography}

\clearpage
\begin{IEEEbiographynophoto}{Taja Kuzman} received the M.S. degree in translation between Slovenian, French and English from the University of Ljubljana, Slovenia, in 2019. Since 2021, she is pursuing her Ph.D. in information and communication technologies at the Jožef Stefan International Postgraduate School, Ljubljana, Slovenia.

Since 2021, she is a Research Assistant at the Department of Knowledge Technologies at the Jožef Stefan Institute in Ljubljana, Slovenia. Her research interests involve language resource collection and curation, and the development of technologies for automatic annotation using deep learning, including text classification tasks, such as automatic genre identification. 

Ms. Kuzman is 
a co-leader of CLASSLA, the CLARIN ERIC knowledge center, which offers expertise on language resources and technologies for South Slavic languages. She is one of the main developers of the CLASSLA-web corpora, the largest text collections for these languages, and conducts evaluations of NLP models on South Slavic languages and dialects. As an author or co-author of 68 resources, she is one of the main contributors of text datasets and language technologies for Slovenian and other European languages at the CLARIN.SI repository.
\end{IEEEbiographynophoto}

\begin{IEEEbiographynophoto}{Nikola Ljube{\v{s}}i{\'c}} received his PhD at the University of Zagreb on the topic of event detection in news data streams in 2009. Currently he works as senior researcher at the Department of Knowledge Technologies at the Jožef Stefan Institute in Ljubljana, Slovenia. He is also affiliated with the Faculty for Computer and Information Science at the University of Ljubljana, as well as with the Institute for Contemporary History in Ljubljana.

His research interests lie in the fields of computational linguistics and computational social science, with an emphasis on South Slavic linguistic and cultural space. He is a co-author of the CLASSLA-Stanza linguistic processing pipeline and CLASSLA-web corpora, both covering the range of South Slavic languages. He also co-authored the training data and various benchmarks for the same language group. His general interest in language variation has motivated him to expand his focus from textual modality to speech modality, inside which he is developing a methodology to scale spoken data collection and automatic enrichment. He is co-leading the CLASSLA knowledge center for South Slavic languages with the first author.

\end{IEEEbiographynophoto}

\EOD

\end{document}